\documentclass{article}
\usepackage{spconf,amsmath,graphicx,hyperref,amssymb,amsfonts}
\usepackage[table]{xcolor}
\usepackage{multirow}
\usepackage{booktabs}
\definecolor{mypurple}{HTML}{e1d5ff}
\newtheorem{observation}{Observation}
\newtheorem{remark}{Remark}
\title{TensorLOB: Structure-Preserving Tensor Learning for Limit Order Book Forecasting}
\name{Yuxuan Gu$^{1}$, Danilo Mandic$^{1}$}
\address{$^{1}$ Department of Electrical and Electronic Engineering,
Imperial College London, United Kingdom}
\begin{document}
\ninept
\maketitle
\begin{abstract}
The limit order book (LOB) records outstanding buy and sell orders across multiple price levels. Its state evolves over time across market side, price level, and order attribute, naturally forming a multiway structure. However, existing LOB models typically flatten these dimensions into a matrix model, thus obscuring the existing structural relationships. Starting from first principles and the market-microstructure intuition of the LOB, we design \textbf{TensorLOB} to operate entirely in the tensor domain. TensorLOB uses mode-wise tensor contractions for feature extraction and directly contracts the learned representation with low-rank class-weight tensors for prediction, retaining the distinct roles of time, side, level, and attribute modes. Experiments on the FI-2010 benchmark show that TensorLOB achieves the highest accuracy, recall, and F1 score across all four evaluated horizons. This proof of concept highlights the potential of high-dimensional modelling of multi-way latent information intrinsic to financial markets, which has so far been largely overlooked.
\end{abstract}
\begin{keywords}
Limit Order Book, Mid-Price Prediction, Tensor Classification, CP Decomposition, Deep Learning
\end{keywords}
\section{Introduction}
The limit order book (LOB) records outstanding buy and sell orders of a traded asset at multiple price levels. Its state evolves whenever orders are submitted, cancelled, or executed, potentially affecting the best bid and ask prices. Predicting short-term movements of the mid-price, which is the average of the best ask and bid prices, is therefore important for market making, execution, and high-frequency risk management. Yet this remains a challenging prediction problem since LOB data are noisy and non-stationary, while the resulting movement classes are often highly imbalanced~\cite{ntakaris2018benchmark}.

At each event, the LOB comprises two sides: ask and bid. Each side is arranged over an ordered sequence of price levels, with price and volume providing different attributes. Across successive events, those changes capture how orders on both sides evolve over time. Hence, a LOB history is naturally a higher-order tensor, with time, market side, price level and attribute as tensor modes.

Despite the naturally multiway structure of LOB, most existing deep LOB prediction methods do not consider the full higher-order structure of the book throughout representation learning. Convolutional, recurrent, attention-based, and bilinear architectures have achieved strong predictive performance~\cite{zhang2019deeplob,zhang2021multi,tran2018temporal,tran2021data}, but operate on vectorized or matricized representations that collapse several economically meaningful LOB dimensions into a single feature axis. Consequently, market side, price level, and order attribute are not retained as distinct modes, and their structural relationships must instead be learned implicitly from data, yielding a weaker inductive bias for LOB modelling. This motivates a prediction framework that preserves and exploits the native multiway structure of the LOB throughout representation learning, as the subject of this work.

\begin{figure}
    \centering
    \includegraphics[width=0.8\linewidth]{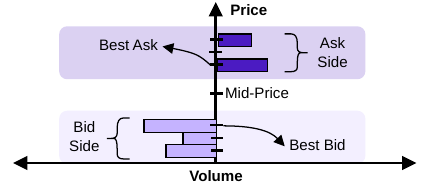}
    \caption{A snapshot of a limit order book snapshot at a single event.}
    \label{fig:lob_diagram}
\end{figure}

To this end, we propose \textbf{TensorLOB}, a structure-preserving neural model that operates directly on the tensor representation of LOB data. TensorLOB uses the native LOB modes to guide representation learning throughout the network. Mode-aware normalization captures temporal and cross-sectional variation, while multi-head Kronecker-factorized tensor projections learn mode-specific interactions without collapsing the side and level axes. Finally, the tensor classifier combines information across time, side, level and learned features for mid-price movement prediction.

The proposed structure-preserving design is evaluated over comprehensive experiments against the state-of-the-art (SOTA) on the FI-2010 benchmark~\cite{ntakaris2018benchmark}. Our proposed TensorLOB achieves the highest mean accuracy, weighted recall, and weighted F1 across a range of prediction horizons. Such consistent performance gains indicate that explicitly preserving the native LOB structure provides an effective inductive bias for mid-price movement prediction. 

\begin{figure*}[h]
    \centering
    \includegraphics[width=\linewidth]{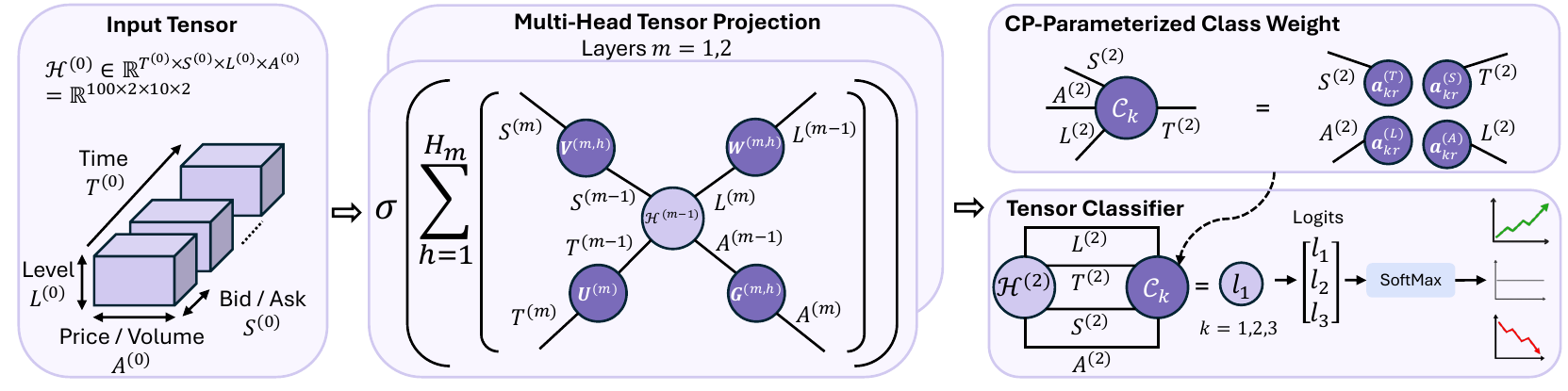}
    \caption{An overview of the TensorLOB architecture with multi-head tensor projection layers and a tensor classifier.}
    \label{fig:architecture}
\end{figure*}

In summary, our main contributions are threefold:
\begin{itemize}
    \item We derive TensorLOB from first-principles market-microstructure intuition, representing the LOB through its native time, side, level, and attribute modes and performing the prediction pipeline entirely in the tensor domain.
    
    \item We develop a structure-preserving architecture based on mode-wise tensor contractions for feature extraction and low-rank class-weight tensors for prediction, enabling mode-specific interactions while retaining the native multiway structure of the LOB.

    \item We conduct extensive experiments on the FI-2010 benchmark, demonstrating consistent improvements in accuracy and weighted F1 across multiple prediction horizons over vectorized, bilinear, recurrent, and conventional baselines.
\end{itemize}

\section{Related Work}
\subsection{LOB Movement Prediction}\label{sec:related_work}
DeepLOB~\cite{zhang2019deeplob} combines convolutional neural networks (CNNs) for feature extraction from vectorized LOB inputs with an LSTM for temporal modelling. DeepLOB-Seq2Seq and DeepLOB-Attention~\cite{zhang2021multi} extend this architecture to multi-horizon prediction through sequence-to-sequence and attention-based decoding, respectively. An alternative line of work is based on temporal attention-augmented bilinear networks (TABL)~\cite{tran2018temporal}. B(TABL) applies bilinear transformations along the flattened feature and temporal axes, while C(TABL) further introduces temporal attention to learn importance of different time steps. BiN-C(TABL)~\cite{tran2021data} augments this framework with learnable normalization along both flattened feature and temporal axes to account for different input scales. These methods improve feature extraction, temporal modelling, and input normalization, but represent the LOB data as a vector or a matrix, with market side, price level, and any order attribute collapsed to a single feature dimension.

\subsection{Tensor Methods}
Tensor methods have been widely used in various ways with neural networks, such as large language model (LLM) compression~\cite{xu2023tensorgpt, Tensorllm, luo2025trawl}, parameter-efficient finetuning~\cite{yang2024loretta, gu2026tera}, tensorized neural representation~\cite{tngram, Kromhc}, and interpretable financial prediction~\cite{xu2023graph, xu2021tensor}. In LOB prediction, Roa-Vicens \textit{et al.}~\cite{roa2022graph} embedded a multi-linear graph filter into tensor-train recurrent networks for high-dimensional LOB data. In contrast, TensorLOB operates directly on the native LOB tensor throughout the network during both feature extraction and classification stages. This provides a more direct structure-preserving inductive bias while remaining parameter-efficient.

\section{Methodology}

\subsection{Preliminaries}
\subsubsection{Limit Order Book}
A limit order book (LOB) is a collection of outstanding buy and sell orders of a financial instrument at discrete price levels. As Fig. \ref{fig:lob_diagram} shows, the ask side contains sell orders with prices greater than the mid-price, while the bid side contains buy orders with prices lower than the mid-price. The best ask and bid prices are the lowest ask price and the highest bid price, respectively, and their average defines the mid-price. On each side, there are multiple levels of orders, with each level characterized by its price and volume.

\subsubsection{Tensor Decomposition}

A tensor is a multi-dimensional array whose dimensions are referred to as
modes. Following~\cite{cichocki2015tensor}, we use $a$, $\mathbf{a}$, $\mathbf{A}$, and $\mathcal{A}$ to denote scalars, vectors, matrices, and tensors, respectively. For an order-$N$ tensor
$\mathcal{A}\in\mathbb{R}^{I_1\times\cdots\times I_N}$, its
$(i_1,\ldots,i_N)$th entry is denoted by $a_{i_1\cdots i_N}$.

The \textbf{mode-$n$ product} of an $N$th-order tensor
$\mathcal{A}\in\mathbb{R}^{I_1\times\cdots\times I_N}$ with a matrix
$\mathbf{M}\in\mathbb{R}^{J\times I_n}$ is denoted by
$\mathcal{A}\times_n\mathbf{M}$. The resulting tensor has shape
$I_1\times\cdots\times I_{n-1}\times J\times I_{n+1}\times\cdots\times I_N$
and is defined elementwise as
\begin{equation}
\left(\mathcal{A}\times_n\mathbf{M}\right)
_{i_1,\ldots,i_{n-1},j,i_{n+1},\ldots,i_N}
=
\sum_{q=1}^{I_n}
a_{i_1,\ldots,i_{n-1},q,i_{n+1},\ldots,i_N}m_{jq}.
\label{eq:mode_product}
\end{equation}

More generally, let
$\mathcal{A}\in\mathbb{R}^{I_1\times\cdots\times I_N}$ and
$\mathcal{B}\in\mathbb{R}^{J_1\times\cdots\times J_M}$. Their contraction over
$P$ pairs of modes $(i_p,j_p)$, where $I_{i_p}=J_{j_p}$ for
$p=1,\ldots,P$, is denoted by $\mathcal{A}
\times_{i_1,\ldots,i_P}^{j_1,\ldots,j_P}
\mathcal{B}$. The elementwise definition is given by 
\begin{equation}
    c_{\mathbf{i}_{\bar{\alpha}}\mathbf{j}_{\bar{\beta}}}
=
\sum_{q_1=1}^{I_{\alpha_1}}
\cdots
\sum_{q_P=1}^{I_{\alpha_P}}
a_{\mathbf{i}_{\bar{\alpha}},q_1,\ldots,q_P}\,
b_{\mathbf{j}_{\bar{\beta}},q_1,\ldots,q_P},
\end{equation}
where $\mathbf{i}_{\bar{\alpha}}$ and $\mathbf{j}_{\bar{\beta}}$ denote the uncontracted mode indices, respectively.

The \textbf{Canonical Polyadic (CP) Decomposition} represents an order-$N$ tensor
as a sum of rank-one outer products. A rank-$R$ CP factorization of
$\mathcal{A}\in\mathbb{R}^{I_1\times\cdots\times I_N}$ is given by
\begin{equation}
\mathcal{A}
=
\sum_{r=1}^{R}
\mathbf{a}_{r}^{(1)}
\circ
\mathbf{a}_{r}^{(2)}
\circ\cdots\circ
\mathbf{a}_{r}^{(N)},
\label{eq:cp}
\end{equation}
where
$\mathbf{a}_{r}^{(n)}\in\mathbb{R}^{I_n}$ is the $r$-th factor vector
associated with mode $n$. Each term
$\mathbf{a}_{r}^{(1)}\circ\cdots\circ\mathbf{a}_{r}^{(N)}$
is therefore a rank-one tensor of size
$I_1\times\cdots\times I_N$. This representation requires $R\sum_{n=1}^{N} I_n$
factor parameters, compared with $\prod_{n=1}^{N} I_n$ for an unstructured tensor, which is parameter-efficient when $R\sum_n I_n \ll \Pi_n I_n$.

\subsection{TensorLOB Architecture}
TensorLOB takes as input the most recent $T=100$ LOB events, arranged
into a fourth-order tensor $\mathcal{X}\in\mathbb{R}^{T^{(0)}\times S^{(0)}\times L^{(0)}\times A^{(0)}}=\mathbb{R}^{100\times2\times10\times2}$, with the four modes as time $T^{(0)}$, market side (ask or bid) $S^{(0)}$, price
level $L^{(0)}$, and attribute (price or volume) $A^{(0)}$, respectively. Rather than flattening these modes into a single feature dimension, TensorLOB directly operates on the 4-$D$ structure throughout the prediction pipeline. As illustrated in Fig.~\ref{fig:architecture}, the architecture
consists of a mode-aware input normalization, two multi-head tensor projection layers for feature extraction, and a CP-parameterized tensor classifier for movement prediction.

\subsubsection{Input Normalization}

Following the bilinear normalization (BiN) scheme of~\cite{tran2021data},
we normalize each input tensor from two complementary views. The temporal
view normalizes each book entry across the $T$ events, while the
cross-sectional view normalizes the book state within each event across the
non-temporal modes. The two normalized tensors are then combined using
learned non-negative weights, with the normalization statistics computed
independently for each input sample and the affine parameters learned during
training.

\subsubsection{Multi-Head Kronecker-Factorized Feature Extraction}

Let $\mathcal{H}^{(m-1)} \in
\mathbb{R}^{T_{m-1}\times S_{m-1}\times L_{m-1}\times A_{m-1}}$
denote the input to feature extraction layer $m$, with
$\mathcal{H}^{(0)}=\widetilde{\mathcal{X}}$. Each layer
performs feature extraction through mode-wise tensor contractions using $H_m$ projection heads. All heads share
the same transformation along the temporal mode, while each head learns
distinct projections along the remaining modes. This multi-head projection allows the layer to capture complementary mode-specific interaction patterns in parallel, rather than relying on a single projection.

Specifically, these mode-wise projections are parameterized by a temporal factor
matrix, $U^{(m)}\in\mathbb{R}^{T_{m-1}\times T_m}$, together with head-specific factor matrices,
$V^{(m,h)}\in\mathbb{R}^{S_{m-1}\times S_m}$,
$W^{(m,h)}\in\mathbb{R}^{L_{m-1}\times L_m}$, and
$G^{(m,h)}\in\mathbb{R}^{A_{m-1}\times A_m}$, for side, level and attribute modes, respectively. These factor matrices define mode-specific linear transformations while
preserving the tensor structure of the LOB representation. The resulting pre-activation representation at layer $m$ then becomes
\begin{equation}
\begin{aligned}
\mathcal{Q}^{(m)}
=
\frac{1}{\sqrt{H_m}}
\sum_{h=1}^{H_m}
\Bigl(
&\mathcal{H}^{(m-1)}
\times_1^2 (\mathbf{U}^{(m)})^\top
\times_1^2 (\mathbf{V}^{(m,h)})^\top
\\
&\times_1^2 (\mathbf{W}^{(m,h)})^\top
\times_1^2 (\mathbf{G}^{(m,h)})^\top
\Bigr),
\end{aligned}
\label{eq:tensorlob_kronecker}
\end{equation}
where the successive contractions are applied on all modes of $\mathcal{H}^{(m-1)}$, yielding
$\mathcal{Q}^{(m)}\in
\mathbb{R}^{T_m\times S_m\times L_m\times A_m}$. It is worth noting that the tensor formulation in Eq.~(\ref{eq:tensorlob_kronecker}) also admits a
Kronecker-factorized interpretation.

\begin{observation}[Tensor contraction and Kronecker representation]
\label{thm:tucker_kronecker}
Let
$\mathcal{X}\in\mathbb{R}^{I_1\times\cdots\times I_N}$ and let
$\mathbf{A}_n\in\mathbb{R}^{J_n\times I_n}$ for $n=1,\dots,N$. Assume
\begin{equation}
\mathcal{Y}
=
\mathcal{X}
\times_1 \mathbf{A}_1
\times_2 \mathbf{A}_2
\times_3
\cdots
\times_N \mathbf{A}_N .
\end{equation}
Then, the following holds.
\begin{equation}
\operatorname{vec}(\mathcal{Y})
=
\left(
\mathbf{A}_N\otimes \mathbf{A}_{N-1}\otimes\cdots\otimes \mathbf{A}_1
\right)
\operatorname{vec}(\mathcal{X}),
\label{eq:tucker_kronecker_relation}
\end{equation}
where $\otimes$ denotes the Kronecker product.
\end{observation}

Observation~\ref{thm:tucker_kronecker} shows that a sequence of mode-wise tensor
contractions is equivalent to a single Kronecker-factorized linear
transformation after vectorization. Hence, each feature extraction head in
Eq.~(\ref{eq:tensorlob_kronecker}) is equivalent to a
Kronecker-factorized linear transformation after vectorization, while TensorLOB performs the computation directly in the tensor form. The following simple example illustrates how this factorization explicitly encodes mode-wise relationships that are absent from an unrestricted vectorized parameterization.

\begin{remark}\label{ex:modewise_bias}
In Eq.~\eqref{eq:tensorlob_kronecker}, for a head $h$, the
contribution from $\mathcal{H}^{(m-1)}_{tsla}$ to a fixed output entry
$\mathcal{Q}^{(m)}_{kqrf}$ is weighted by
$U^{(m)}_{tk}V^{(m,h)}_{sq}W^{(m,h)}_{lr}G^{(m,h)}_{af}$.
Hence, input entries sharing the same mode index also share the corresponding
factor in their effective weights. For example, two entries at the same
event index $t$ share $U^{(m)}_{tk}$. If they also share the same level $l$,
they additionally share $W^{(m,h)}_{lr}$. In contrast, a dense transformation
after vectorization assigns an independent coefficient to each pair of
flattened indices. Thus, TensorLOB explicitly ties coefficients according to
the multiway structure, rather than requiring such relationships to be
learned from data.
\end{remark}

The projected representation is next normalized along the attribute mode, followed
by a GELU activation~\cite{hendrycks2016gaussian} and
dropout~\cite{hinton2012improving},
\begin{equation}
\mathcal{H}^{(m)}
=
\operatorname{Dropout}\left(
\operatorname{GELU}\left(
\operatorname{LN}_{A_m}
\left(\mathcal{Q}^{(m)}\right)
\right)\right).
\label{eq:tensorlob_layer_output}
\end{equation}

The first feature extraction layer uses $H_1=24$ heads and maps a tensor of
size $100\times2\times10\times2$ to $35\times2\times10\times6$. The second
layer uses $H_2=44$ heads and produces a tensor of size
$20\times2\times10\times12$. Thus, the two layers progressively reduce the
temporal dimension, from $100$ to $12$, while expanding the learned attribute dimension, from $2$ to $10$, with the side and level dimensions preserved.

\subsubsection{CP-Parameterized Tensor Classification}
Given
$\mathcal{H}^{(2)}
\in
\mathbb{R}^{T_2\times S_2\times L_2\times A_2}$,
where $T_2=20$, $S_2=2$, $L_2=10$, and $A_2=12$, instead of first
collapsing all modes into an intermediate vector representation,
we directly map the tensor to the
movement class logits. For each class $k=1,2,3$, let
$\mathcal{C}_k
\in
\mathbb{R}^{T_2\times S_2\times L_2\times A_2}$
denote its classification coefficient tensor. To retain the mode-wise structure also in the prediction layer, we parameterize $\mathcal{C}_k$ using a rank-$R$ CP decomposition,
\begin{equation}
\mathcal{C}_k
=
\sum_{r=1}^{R}
\mathbf{a}^{(T)}_{kr}
\circ
\mathbf{a}^{(S)}_{kr}
\circ
\mathbf{a}^{(L)}_{kr}
\circ
\mathbf{a}^{(A)}_{kr},
\label{eq:tensorlob_cp_classifier_weights}
\end{equation}
where
$\mathbf{a}^{(T)}_{kr}\in\mathbb{R}^{T_2}$,
$\mathbf{a}^{(S)}_{kr}\in\mathbb{R}^{S_2}$,
$\mathbf{a}^{(L)}_{kr}\in\mathbb{R}^{L_2}$, and
$\mathbf{a}^{(A)}_{kr}\in\mathbb{R}^{A_2}$
are the temporal, side, level, and learned-feature factors, respectively.

The logit associated with class $k$ is obtained by contracting
$\mathcal{H}^{(2)}$ with each rank-one CP component along all four modes to yield
\begin{equation}
\ell_k
=
\sum_{r=1}^{R}
\mathcal{H}^{(2)}
\times_{1}
\left(\mathbf{a}^{(T)}_{kr}\right)^{\top}
\times_{2}
\left(\mathbf{a}^{(S)}_{kr}\right)^{\top}
\times_{3}
\left(\mathbf{a}^{(L)}_{kr}\right)^{\top}
\times_{4}
\left(\mathbf{a}^{(A)}_{kr}\right)^{\top}
+
b_k ,
\label{eq:tensorlob_cp_classifier_contraction}
\end{equation}
where $b_k$ is the class-specific bias. Equivalently,
\begin{equation}
\ell_k
=
\left\langle
\mathcal{H}^{(2)},\mathcal{C}_k
\right\rangle
+
b_k .
\label{eq:tensorlob_cp_classifier_inner_product}
\end{equation}
Thus, each rank-one CP component specifies a separable interaction across the modes, while summing over $R$ components allows multiple multilinear interaction patterns to contribute to each class
score. Applying Eq.~(\ref{eq:tensorlob_cp_classifier_contraction}) to all
classes yields the logit vector $\boldsymbol{\ell}=[\ell_1,\ell_2,\ell_3]^T\in \mathbb{R}^{3}$, with the class probabilities are $\mathbf{p} = \operatorname{SoftMax}(\boldsymbol{\ell})$, and the predicted movement, with $\hat{y} = \arg\max_{k\in\{1,2,3\}} p_k$, corresponding to
downward, stationary, and upward movements. 

\begin{remark}
    We use CP rank $R=16$. The CP parameterization requires
$KR(T_2+S_2+L_2+A_2)+K$ learnable classification parameters, compared with
$KT_2S_2L_2A_2+K$ for an unrestricted coefficient tensor. With $K=3,T_2=20, S_2=2, L_2=10$ and $A_2=12$, the CP classifier requires $2{,}115$ parameters, compared with $14{,}403$ for the corresponding dense classifier, representing a 85.3\% parameter reduction in the classification layer.
\end{remark}

\section{Experiments}

\subsection{Experimental Setup}
We evaluated TensorLOB on the pre-normalized FI-2010 benchmark using the provided three-class mid-price movement labels (down, stationary, and up) at prediction horizons of 10, 20, 50, and 100 future events~\cite{ntakaris2018benchmark}. Each sample consists of the most recent 100 LOB events, with the target assigned to the final event of the window. Following the standard setting in the benchmark, the first seven trading days were used for model development and the final three for testing. The development period was split chronologically, with the first 80\% for training and the remaining 20\% for validation. This produced 203,701 training, 50,851 validation, and 139,488 test samples.

All neural models were trained under a common optimization and model-selection criterion. We minimize cross-entropy with label smoothing of 0.05 using AdamW optimizer~\cite{loshchilov2017decoupled}, with an initial learning rate of $2\times10^{-4}$, weight decay of $10^{-4}$, and cosine learning-rate annealing. Training was limited to 60 epochs with a batch size of 1,024, and gradients are clipped to unit norm. Training samples were shuffled within each epoch, whereas the validation and test sets retain their chronological order. The checkpoint with the highest validation weighted-average F1 score was retained, and the training terminates after 12 consecutive epochs without improvement.

We compare TensorLOB with the SOTA neural and bilinear models described in Section~\ref{sec:related_work}, together with conventional machine-learning methods~\cite{tsantekidis2017using}, including multilayer perceptron (MLP), XGBoost, linear support vector machine (SVM), and LSTM. Each experiment was repeated with seeds 7, 42, and 123, and results are reported as the mean $\pm$ sample standard deviation. Performance was measured by accuracy and weighted-average precision, recall, and F1 over the three movement classes. Note that for single-label multiclass classification, weighted average recall is algebraically equal to accuracy, and both were retained for consistency.

\begin{table}[ht!]
\centering
\caption{Comparison of test performance on FI-2010 benchmark of all methods over four prediction horizons $h$. Results are reported as mean $\pm$ sample standard deviation over three random seeds, with the best value for each metric in bold.}
\label{tab:results}
\setlength{\tabcolsep}{0.8pt}
\renewcommand{\arraystretch}{0.85}
\begin{tabular}{cccccc}
\toprule
$h$ & Model & Accuracy $\uparrow$ & Precision $\uparrow$ & Recall $\uparrow$ & F1 $\uparrow$ \\
\midrule

\multirow{12}{*}{10}
 & DeepLOB       & 80.4 $\pm$ 0.9 & 79.0 $\pm$ 0.9 & 80.4 $\pm$ 0.9 & 79.2 $\pm$ 0.8 \\
 & DeepLOB-Seq   & 70.7 $\pm$ 0.0 & 50.0 $\pm$ 0.0 & 70.7 $\pm$ 0.0 & 58.6 $\pm$ 0.0 \\
 & DeepLOB-Att.  & 67.5 $\pm$ 9.1 & 63.6 $\pm$ 7.5 & 67.5 $\pm$ 9.1 & 64.4 $\pm$ 7.8 \\
 & Linear SVM    & 70.7 $\pm$ 0.0 & 50.0 $\pm$ 0.0 & 70.7 $\pm$ 0.0 & 58.6 $\pm$ 0.0 \\
 & MLP           & 70.7 $\pm$ 0.0 & 50.0 $\pm$ 0.0 & 70.7 $\pm$ 0.0 & 58.6 $\pm$ 0.0 \\
 & LSTM          & 70.7 $\pm$ 0.0 & 50.0 $\pm$ 0.0 & 70.7 $\pm$ 0.0 & 58.6 $\pm$ 0.0 \\
 & B(TABL)       & 70.7 $\pm$ 0.0 & 50.0 $\pm$ 0.0 & 70.7 $\pm$ 0.0 & 58.6 $\pm$ 0.0 \\
 & C(TABL)       & 70.7 $\pm$ 0.0 & 50.0 $\pm$ 0.0 & 70.7 $\pm$ 0.0 & 58.6 $\pm$ 0.0 \\
 & BiN-C(TABL)   & 82.2 $\pm$ 1.7 & 81.7 $\pm$ 2.1 & 82.2 $\pm$ 1.7 & 80.9 $\pm$ 1.7 \\
 & XGBoost       & 78.9 $\pm$ 0.0 & 78.9 $\pm$ 0.0 & 78.9 $\pm$ 0.0 & 75.4 $\pm$ 0.0 \\
 \rowcolor{mypurple}
 & \textbf{TensorLOB}     & \textbf{83.5 $\pm$ 0.1} & \textbf{83.0 $\pm$ 0.1} & \textbf{83.5 $\pm$ 0.1} & \textbf{82.5 $\pm$ 0.1} \\

\midrule

\multirow{12}{*}{20}
 & DeepLOB       & 64.1 $\pm$ 0.8 & 62.8 $\pm$ 1.0 & 64.1 $\pm$ 0.8 & 63.1 $\pm$ 0.7 \\
 & DeepLOB-Seq   & 62.1 $\pm$ 0.0 & 38.5 $\pm$ 0.0 & 62.1 $\pm$ 0.0 & 47.6 $\pm$ 0.0 \\
 & DeepLOB-Att.  & 56.4 $\pm$ 9.9 & 40.8 $\pm$ 3.9 & 56.4 $\pm$ 9.9 & 46.2 $\pm$ 2.3 \\
 & Linear SVM    & 57.5 $\pm$ 1.5 & 46.4 $\pm$ 1.1 & 57.5 $\pm$ 1.5 & 48.5 $\pm$ 0.4 \\
 & MLP           & 62.1 $\pm$ 0.0 & 38.5 $\pm$ 0.0 & 62.1 $\pm$ 0.0 & 47.6 $\pm$ 0.0 \\
 & LSTM          & 62.1 $\pm$ 0.0 & 38.5 $\pm$ 0.0 & 62.1 $\pm$ 0.0 & 47.6 $\pm$ 0.0 \\
 & B(TABL)       & 62.1 $\pm$ 0.0 & 38.5 $\pm$ 0.0 & 62.1 $\pm$ 0.0 & 47.6 $\pm$ 0.0 \\
 & C(TABL)       & 62.1 $\pm$ 0.0 & 38.5 $\pm$ 0.0 & 62.1 $\pm$ 0.0 & 47.6 $\pm$ 0.0 \\
 & BiN-C(TABL)   & 71.4 $\pm$ 4.3 & 70.2 $\pm$ 5.6 & 71.4 $\pm$ 4.3 & 68.9 $\pm$ 4.6 \\
 & XGBoost       & 70.8 $\pm$ 0.0 & 70.9 $\pm$ 0.0 & 70.8 $\pm$ 0.0 & 66.1 $\pm$ 0.0 \\
 \rowcolor{mypurple}
 & \textbf{TensorLOB}  & \textbf{74.0 $\pm$ 0.1} & \textbf{73.2 $\pm$ 0.1} & \textbf{74.0 $\pm$ 0.1} & \textbf{72.2 $\pm$ 0.0} \\



\midrule

\multirow{12}{*}{50}
 & DeepLOB       & 71.7 $\pm$ 1.2 & 71.7 $\pm$ 0.8 & 71.7 $\pm$ 1.2 & 71.6 $\pm$ 0.9 \\
 & DeepLOB-Seq   & 44.4 $\pm$ 12.1 & 42.9 $\pm$ 12.8 & 44.4 $\pm$ 12.1 & 42.4 $\pm$ 12.9 \\
 & DeepLOB-Att.  & 51.1 $\pm$ 5.5 & 47.0 $\pm$ 9.9 & 51.1 $\pm$ 5.5 & 48.5 $\pm$ 7.5 \\
 & Linear SVM    & 43.6 $\pm$ 1.2 & 49.8 $\pm$ 2.8 & 43.6 $\pm$ 1.2 & 42.1 $\pm$ 3.5 \\
 & MLP           & 44.3 $\pm$ 3.3 & 45.1 $\pm$ 2.8 & 44.3 $\pm$ 3.3 & 44.2 $\pm$ 3.5 \\
 & LSTM          & 52.2 $\pm$ 0.4 & 51.9 $\pm$ 0.0 & 52.2 $\pm$ 0.4 & 51.9 $\pm$ 0.1 \\
 & B(TABL)       & 54.2 $\pm$ 7.7 & 53.6 $\pm$ 7.3 & 54.2 $\pm$ 7.7 & 53.7 $\pm$ 7.4 \\
 & C(TABL)       & 63.8 $\pm$ 0.5 & 63.1 $\pm$ 0.6 & 63.8 $\pm$ 0.5 & 63.1 $\pm$ 0.8 \\
 & BiN-C(TABL)   & \textbf{77.5 $\pm$ 0.4} & \textbf{78.0 $\pm$ 0.3} & \textbf{77.5 $\pm$ 0.4} & 77.3 $\pm$ 0.4 \\
 & XGBoost       & 69.8 $\pm$ 0.0 & 69.4 $\pm$ 0.0 & 69.8 $\pm$ 0.0 & 69.2 $\pm$ 0.0 \\
 \rowcolor{mypurple}
 & \textbf{TensorLOB}     & \textbf{77.5 $\pm$ 0.1} & 77.5 $\pm$ 0.1 & \textbf{77.5 $\pm$ 0.1} & \textbf{77.4 $\pm$ 0.1} \\

\midrule

\multirow{12}{*}{100}
 & DeepLOB       & 69.7 $\pm$ 0.5 & 69.9 $\pm$ 0.5 & 69.7 $\pm$ 0.5 & 69.6 $\pm$ 0.5 \\
 & DeepLOB-Seq   & 48.5 $\pm$ 4.4 & 49.4 $\pm$ 4.3 & 48.5 $\pm$ 4.4 & 48.1 $\pm$ 5.1 \\
 & DeepLOB-Att.  & 47.5 $\pm$ 10.4 & 46.8 $\pm$ 19.7 & 47.5 $\pm$ 10.4 & 42.8 $\pm$ 12.6 \\
 & Linear SVM    & 42.0 $\pm$ 0.4 & 43.2 $\pm$ 0.1 & 42.0 $\pm$ 0.4 & 39.2 $\pm$ 1.1 \\
 & MLP           & 43.9 $\pm$ 0.4 & 49.4 $\pm$ 0.6 & 43.9 $\pm$ 0.4 & 38.9 $\pm$ 1.1 \\
 & LSTM          & 47.7 $\pm$ 0.5 & 48.1 $\pm$ 0.9 & 47.7 $\pm$ 0.5 & 47.4 $\pm$ 0.6 \\
 & B(TABL)       & 48.2 $\pm$ 4.2 & 51.4 $\pm$ 5.7 & 48.2 $\pm$ 4.2 & 47.6 $\pm$ 4.1 \\
 & C(TABL)       & 63.2 $\pm$ 2.0 & 63.7 $\pm$ 1.7 & 63.2 $\pm$ 2.0 & 63.1 $\pm$ 2.2 \\
 & BiN-C(TABL)   & 77.9 $\pm$ 0.2 & 78.0 $\pm$ 0.2 & 77.9 $\pm$ 0.2 & 78.0 $\pm$ 0.2 \\
 & XGBoost       & 61.2 $\pm$ 0.0 & 65.3 $\pm$ 0.0 & 61.2 $\pm$ 0.0 & 60.8 $\pm$ 0.0 \\
\rowcolor{mypurple}
 & \textbf{TensorLOB}     & \textbf{79.1 $\pm$ 0.8} & \textbf{79.1 $\pm$ 0.8} & \textbf{79.1 $\pm$ 0.8} & \textbf{79.1 $\pm$ 0.8} \\
\bottomrule
\end{tabular}
\end{table}

\subsection{Numerical Results}

Table~\ref{tab:results} shows that the advantage of TensorLOB is not limited to a particular metric or prediction horizon. TensorLOB achieves the highest mean accuracy, weighted recall, and weighted F1 across all four prediction horizons, while also attaining the highest weighted precision at three of them. The improvement is most pronounced at the 20-event horizon, where its weighted F1 exceeds that of the secong best method, BiN-C(TABL), by 3.3 percentage points. Although the gap becomes smaller at horizon 50, TensorLOB retains the lead and the margin widens again at horizon 100. More broadly, its gains over DeepLOB become particularly substantial at the 20- and 100-event horizons, indicating that the advantage persists as the prediction task changes in temporal scale.

These results are consistent with the benefit of the structure-preserving design of TensorLOB. While several baselines deteriorate considerably as the horizon changes, and some collapse to predicting only the majority stationary class at horizons 10 and 20, TensorLOB maintains strong performance across all considered metrics. By retaining the time, side, level, and attribute modes, the model is able to exploit mode-specific information that is obscured by flattening through vectorization or matricization as in the current models. In particular, the mode-wise tensor contractions provide structured feature interactions, while the tensor classifier directly combines these interactions for classification without discarding the underlying LOB geometry. The consistent gains over the vectorized baselines support tensor structure as an effective inductive bias for LOB prediction.

\section{Conclusion}
We have introduced TensorLOB, a tensor structure-preserving model for limit-order-book prediction. Starting from first principles and domain knowledge of LOB mechanics, TensorLOB retains the natural time, side, level, and attribute modes and performs representation learning and prediction entirely in the tensor domain. By combining mode-specific tensor contractions with tensor classification, the model exploits the underlying multi-way structure while remaining parameter efficient. Across a range of FI-2010 prediction horizons and three random seeds, TensorLOB achieved the highest mean accuracy and weighted F1. As a proof of concept, these results opens up new perspective in financial modelling by resorting to the high-dimensional modelling of multi-way latent information, native to financial markets, which has so far been largely overlooked.

\section{Compliance with Ethical Standards}
This is a numerical simulation study for which no ethical approval was required.

\bibliographystyle{IEEEbib}
\bibliography{refs}

\end{document}